\pdfoutput=1
\documentclass[11pt]{article}

\usepackage{EMNLP2022}

\usepackage{times}
\usepackage{latexsym}

\usepackage[T1]{fontenc}
\usepackage[utf8]{inputenc}

\usepackage{microtype}

\usepackage{inconsolata}

\usepackage{graphicx}
\usepackage{tabularx}
\usepackage{amsmath}
\usepackage{amssymb}
\usepackage{multirow}
\title{Towards Explaining Subjective Ground of Individuals on Social Media}

\author{Younghun Lee \and Dan Goldwasser \\
       Department of Computer Science\\
        Purdue University\\
        West Lafayette, IN, USA\\
        \texttt{\{younghun, dgoldwas\}@purdue.edu} \\}

\begin{document}
\maketitle
\begin{abstract}
Large-scale language models have been reducing the gap between machines and humans in understanding the real world, yet understanding an individual's theory of mind and behavior from text is far from being resolved. 

This research proposes a neural model---Subjective Ground Attention---that learns subjective grounds of individuals and accounts for their judgments on situations of others posted on social media. Using simple attention modules as well as taking one's previous activities into consideration, we empirically show that our model provides human-readable explanations of an individual's subjective preference in judging social situations. We further qualitatively evaluate the explanations generated by the model and claim that our model learns an individual's subjective orientation towards abstract moral concepts.%more consistently compared to other baseline models.
\end{abstract}

\section{Introduction}

For the last few years, large-scale language models have shown substantial performance gains in many different sub-fields of natural language processing \citep{liu2019roberta, raffel2020exploring}. Researchers have hypothesized that such language models contain knowledge of linguistic characteristics, logical inference, and real world events in their parameters, and this knowledge can be fine-tuned and adapted to downstream tasks \citep{wang2019superglue}. The recent success of commonsense reasoning, for instance, shows that language model parameters can be used as a knowledge base while they comprehensively learn commonsense patterns \citep{hwang2021comet}. 

Although deeper and larger language models have led machines to better comprehend how the real world works, understanding an individual’s perspective and behavior from text is yet far from being resolved. Humans perceive daily situations and events differently, and employ certain biases \citep{kahneman2011thinking} and social expectations \citep{hilton1990conversational} when they evaluate the given event and social behavior of others \citep{miller2019explanation}. An individual's process of attributing, evaluating, and explaining an event has been widely investigated by cognitive and social psychologists for the past few decades \citep{mcclure2002goal, hilton2017social}, yet its application to neural language models is outside the mainstream natural language processing research.

\begin{figure}
    \centering
    \includegraphics[width=0.45\textwidth]{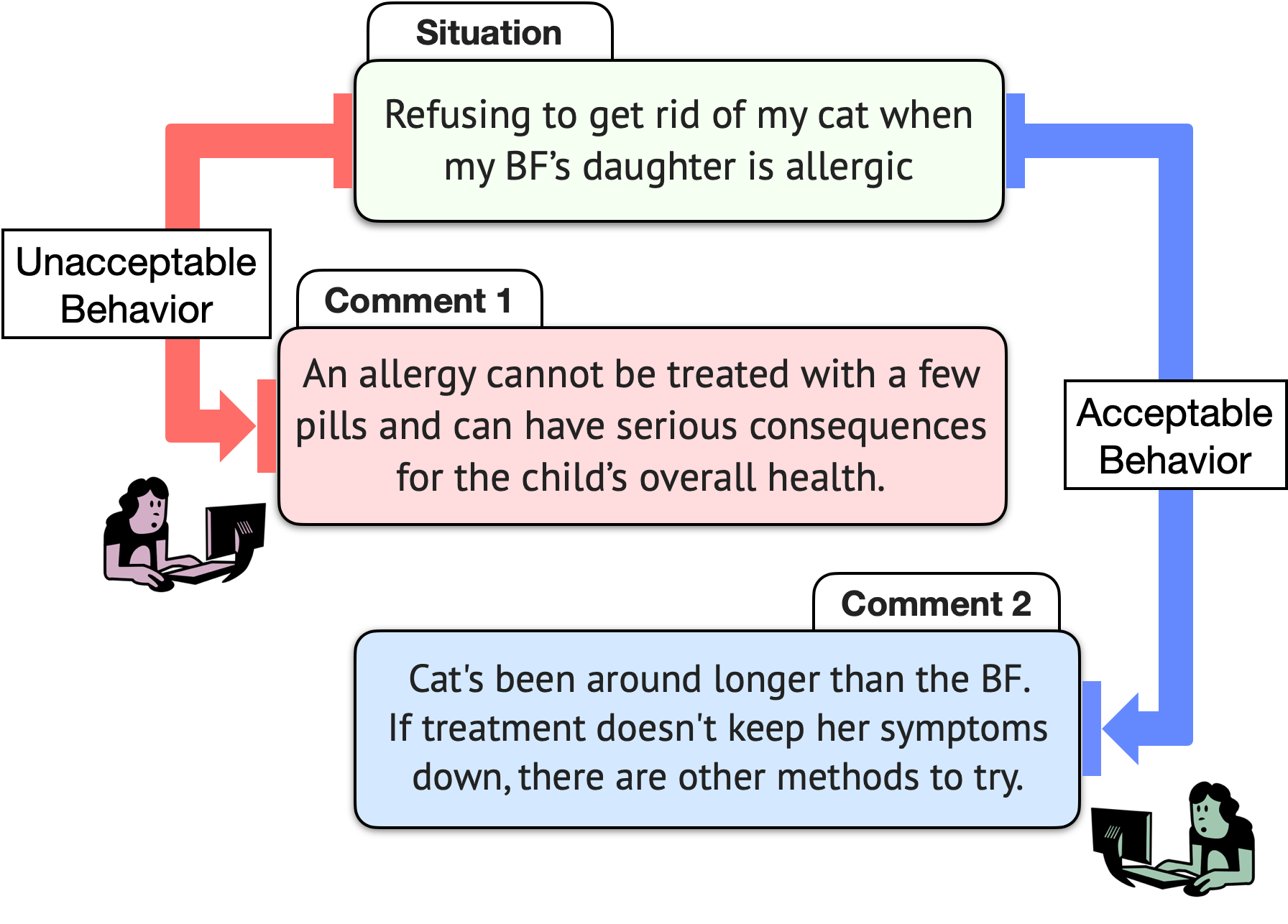}
    \caption{An example of a reddit post and its comments crawled from \texttt{r/AmITheAsshole}. The author describes a situation of not giving up on the pet over the health of significant other's daughter. The author's behavior is unacceptable to the first redditor (red arrow), while the second redditor (blue arrow) has an opposite view.}
    \label{fig:example-intro}
\end{figure}
%DG: add more details about how the task is formulated; how subjective ground is modeled computationally (past situations etc), abstract concepts analysis
This paper proposes a neural model, Subjective Ground Attention, that learns cognitive models of individuals and explains their subjective judgments on situations that are posted on social media. We analyze a Reddit community, \verb|r/AmITheAsshole|, where users submit posts asking whether their behaviors are justifiable, and other users leave comments with their judgments. Figure \ref{fig:example-intro} shows an example of a reddit post where an author describes a situation, and different redditors provide their subjective judgment through comments.

The research hypothesis is that people comprehend and account for others' situations based on their subjective ground, a maxim that plays a central role in human moral judgments \citep{neuhouser1990fichte}, which can be represented from their previous activities. To investigate this hypothesis, we formulate a task to predict the redditors' judgment given diverse situations. Each redditor is represented by their subjective ground which is estimated from their previous comments. Using clustering methods, the model selects a set of the most relevant subjective ground to a given situation. The model then learns attention weights among subjective ground comments to predict the redditor's moral judgments. Through learned attention weights, we present human-readable subjective ground and how they contribute to the model's prediction about the judgments. 
%The research hypothesis is that people comprehend and account for others' situations based on their subjective ground, a maxim that plays a central role in human moral judgments \citep{neuhouser1990fichte}, which can be estimated from their previous activities. % For better explainability of the model, 

From empirical results, we suggest that our proposed model provides explanations for the redditors' subjective judgments on diverse social situations and contributes to downstream task performance in a statistically significant way. We additionally compute the consistency score of attention weights with respect to the model's final prediction, and show that the model efficiently uses attended subjective ground. From qualitative analysis results, we further claim that our model not only learns an individual's subjective preference on real world situations (e.g. reporting my best friend for cheating), but also infers an individual's perspective on more abstract concepts of competing moral values (e.g. fairness is more important than a friendship).
%we empirically show that our model provides explanations for the redditor judgments on diverse social situations with more accurate and consistent predictions. 

\textbf{Key Contributions}: To the best of our knowledge, this is the first attempt to estimate subjective ground of individuals, using it to explain their activities on social media. With better representation of human cognitive models and real world situations, we expect machines to perform more meaningful and accurate inference. This would ultimately help artificial intelligence agents by enabling maximal personalization; not only will it remember an individual’s previous history and preference, it will empathize with one's state and situation in a human-understandable way.
%DG: add more details about how the task is formulated; how subjective ground is modeled computationally (past situations etc), abstract concepts analysis

\begin{table}\footnotesize
\centering
\begin{tabularx}{0.98\columnwidth}{l|r}
\hline
\multicolumn{2}{c}{Modified  \texttt{ Social Chemistry 101} ($\mathcal{D}$)}\\
\hline
\# of total instances & 14,391 \\
\# of unique situations & 9,663 \\
Max / Min \# of instances per redditor & 965 / 298\\
\# Acceptable / Unacceptable labels & 9,817 / 4,574\\
\hline
\hline
\multicolumn{2}{c}{Crawled \texttt{ r/AmITheAsshole} ($\mathcal{D}^+$)}\\
\hline
\# of total instances & 66,603 \\
\# of unique situations & 52,075 \\
Max / Min \# of instances per redditor & 9,711 / 513\\
\# Acceptable / Unacceptable labels & 42,961 / 23,642\\
\hline
\end{tabularx}
\caption{\label{tab:data-statistics}
Statistics of the two datasets. Both datasets take the most active 30 redditors into consideration, keeping the instances that contain coded judgments in the comments.
}
\end{table}

\section{Data Preparation}
We analyze daily situations and individuals' subjective judgments on them posted on a Reddit community, \verb|r/AmITheAsshole|. Users submit posts describing their situations and ask whether or not their behaviors are acceptable. One of the advantages of using this data domain is that most of the situations are generic (e.g. getting annoyed at my roommate) rather than related to specific world events (e.g. new climate change policies in U.S.), thus the models benefit from implicit knowledge in language model parameters to better understand the situation. %thus the models do not require event relations or world knowledge to fully understand the situation.

\subsection{\texttt{Social Chemistry 101} Dataset}
\citet{forbes-etal-2020-social} annotated moral rules-of-thumb (RoT) that can be used in judging whether or not the input situations are acceptable. The authors released \verb|Social Chemistry 101| dataset, which contains around 30k situations posted on \verb|r/AmITheAsshole|. Consider the following situation and its rules-of-thumb as an example:
\begin{quote}\footnotesize
\textbf{Situation}: Asking someone at the gym to stop talking to me.

\textbf{RoT 1}: It is okay to not want to randomly make new friends.\\
\textbf{RoT 2}: It is expected that you are kind when others are extroverted and try to speak to you.
\end{quote}

\begin{figure*}[t!]
    \centering
    \includegraphics[width=0.88\textwidth]{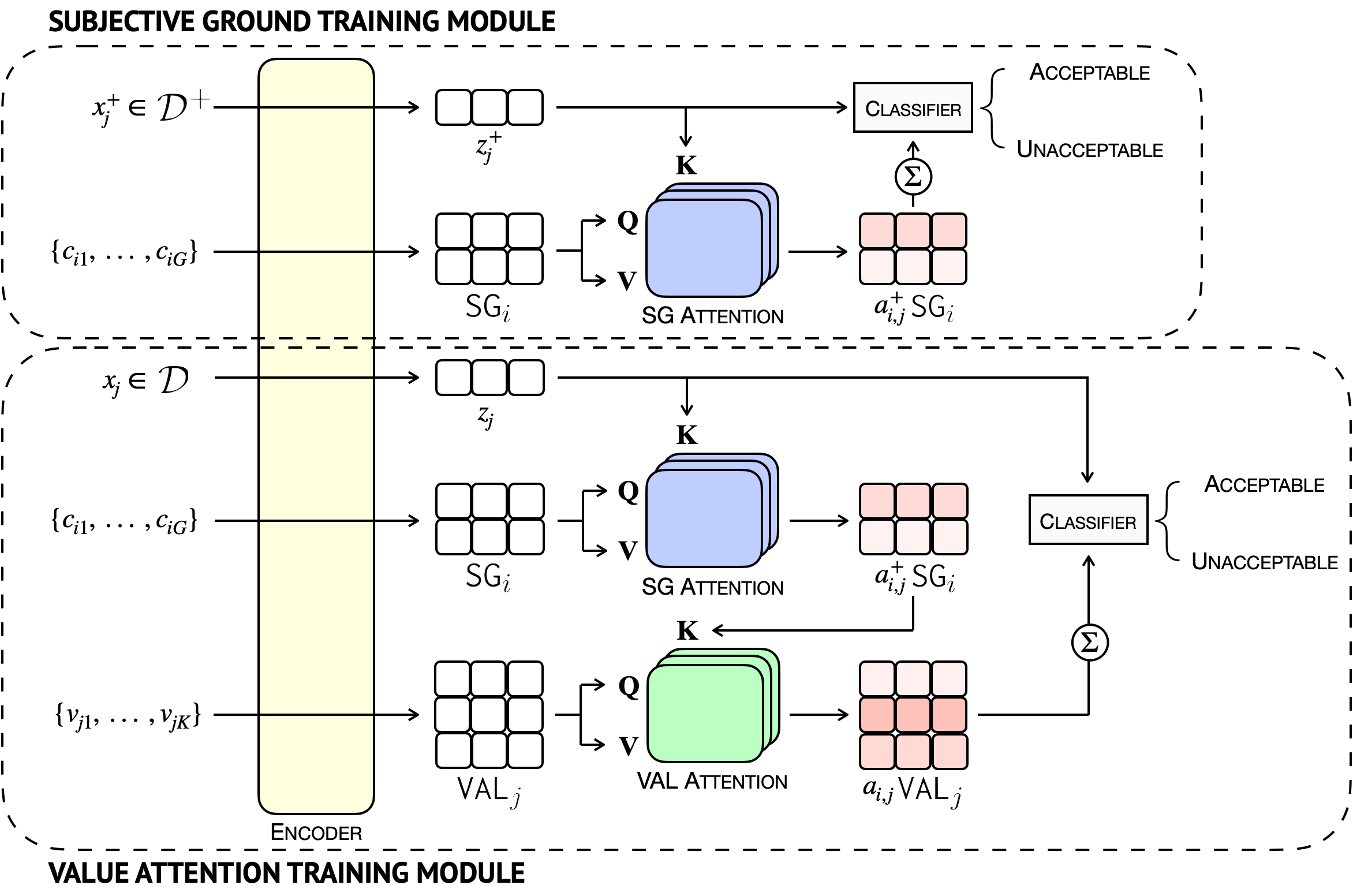}
    \caption{Training process of our proposed model. $x$ is input situations, $c$ is subjective ground comments, and $v$ is rule-of-thumb candidates. After the subjective ground module has been trained on $\mathcal{D}^+$, the parameters of the encoder and the subjective ground attention layers are shared in value attention training.}
    \label{fig:model}
\end{figure*}

% \subsection{Extending Rules-of-Thumb}
We make use of moral rule-of-thumb annotations as a tool for explaining an individual's subjectivity. Observing the number of annotated rules-of-thumb is small for many instances and most of these instances have rules-of-thumb supporting only one side of moral judgment, we manually extend rules-of-thumb. Each rule-of-thumb annotation in \verb|Social Chemistry 101| consists of a judgment (e.g. It is okay) and an action (e.g. not wanting to randomly make new friends). We extend the rules-of-thumb for each situation by replacing judgment words (e.g. It is okay $\rightarrow$ It is not okay) while keeping the action description. We prioritize replacing judgment words with their opposite meaning, which is crucial to ensure obtaining both sides of rules-of-thumb. To efficiently train the model, we set a fixed number, 5, as the number of rules-of-thumb for all situations.
%Because the number of annotated rules-of-thumb is small for many instances, we extend it by manually replacing judgment words.

% \subsection{Identifying Active Redditors}
A small number of training instances prevent the model from generalization. We thus identify redditors who are actively involved in \verb|r/AmITheAsshole|; we focus on the 30 redditors who have commented the most on the posts in the \verb|Social Chemistry 101| dataset.
%One of the main components of this research is to model the subjectivity of individuals. As a small number of training instances prevent the model from generalization, we identify redditors who have actively commented on posts in the \verb|Social Chemistry 101| dataset. We obtain the list of 30 most active redditors who have commented on more than 300 posts and crawl their comments, denoting the dataset as $\mathcal{D}$. 

\subsection{Crawling from \texttt{r/AmITheAsshole}}
In this work, we estimate an individual's subjective ground using their previous activities. The redditor's previous activities are defined as the comments they have left on \verb|r/AmITheAsshole|; comments on other subreddits are mostly irrelevant to moral judgments. We crawl active redditors' comments\footnote{We mainly used Pushshift Reddit API (https://github.com/ pushshift/api) and Python Reddit API Wrapper (https://praw. readthedocs.io/).} and denote the intersection of crawled data and \verb|Social Chemistry 101| as $\mathcal{D}$. All other instances of the crawled data are denoted as $\mathcal{D}^+$. Note that the instances in $\mathcal{D}$ have annotated rules-of-thumb while $\mathcal{D}^+$ doesn't.

% Using Pushshift Reddit API\footnote{https://github.com/pushshift/api} and Python Reddit API Wrapper (PRAW)\footnote{https://praw.readthedocs.io/}, we crawl \textcolor{red}{active redditors' comments left on} \verb|r/AmITheAsshole|. \textcolor{red}{We denote the intersection of crawled data and} \verb|Social Chemistry 101| as $\mathcal{D}$. 
%to crawl posts and comments. We denote this additionally crawled data as $\mathcal{D}^+$.

\subsection{Preprocess Comments and Obtain Moral Judgments}
Rather than predicting the authorship of the comments, this work solely focuses on analyzing the moral judgment. This is to prevent the model from picking up shallow features, such as a redditor's linguistic styles, without focusing on learning their subjectivity. 

We preprocess the redditors' comments and obtain their moral judgments on input situations. In the \verb|r/AmITheAsshole| community, redditors provide their judgments on the situation with pre-defined codes; \verb|YTA| (You're The Asshole), \verb|NTA| (Not The A-hole), \verb|ESH| (Everyone Sucks Here), \verb|NAH| (No A-holes Here), and \verb|INFO| (Not Enough Info).\footnote{Abbreviation for the moral judgments is described in https://www.reddit.com/r/AmItheAsshole/} We identify these code words from the comments and mark them as the redditor's judgments on the situation. We group \verb|NTA| and \verb|NAH| as `acceptable', and \verb|YTA| and \verb|ESH| as `unacceptable'. Instances with the code \verb|INFO| are discarded, as there is no moral subjectivity included in them. 
Detailed statistics of the two datasets are described in Table \ref{tab:data-statistics}.

\begin{table*}
    \centering
    \includegraphics[width=0.98\textwidth]{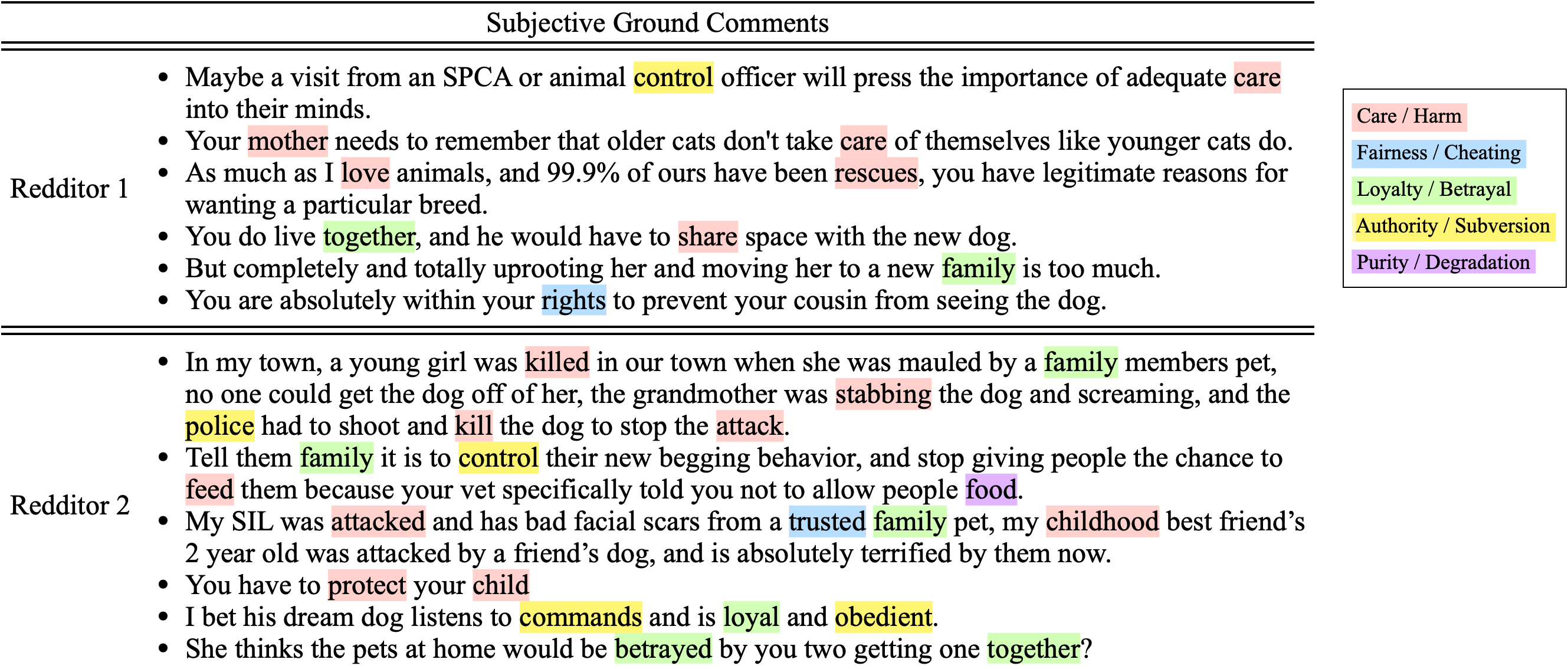}
    \caption{Subjective ground comments of the two redditors in the same cluster. The topic of this cluster is pet / companion animals, and the two redditors express different subjective ground---one mentions methods providing better environments for the pets (Redditor 1), while the other feels that pets can be harmful (Redditor 2). Color-coded parts in each comment indicate the words that match with the Moral Foundation Dictionary vocabularies.}
    \label{tab:mg-example}
\end{table*}

\section{Model}

In this work, we develop neural models with two main components: subjective ground training module and value attention training module. This is followed by a classifier to predict the redditor's moral judgments. Figure \ref{fig:model} illustrates the model diagram. Mathematical details of the model components are described in Appendix \ref{sec:model-formalization}

\subsection{Subjective Ground Training Module}
Subjective ground base consists of a set of previous comments left on \verb|r/AmITheAsshole|. We hypothesize that an individual's subjectivity towards situations related to a specific topic can be applied to other situations within the same topic. For instance, if an individual has a positive subjectivity in raising pets, their moral judgments on animal abuse would be `unacceptable'. Following this intuition, we vectorize input situations in $\mathcal{D}^+$ using Sentence-BERT \citep{reimers-gurevych-2019-sentence}, apply K-Means clustering to identify a fixed number of topic groups among situations, and cluster the redditor's comments on situations within the same topic group. We set the number of clusters to 20, based on the computed intertia values. 

Recognizing the majority of comments in the subjective ground base are not informative with respect to estimating the redditor's subjectivity, we prune unnecessary comments and obtain the subjective ground that is potentially more relevant. In order to determine the relevance of subjective ground comments, we apply Moral Foundations Theory \citep{haidt2004intuitive, graham2013moral}, a framework to explain the origins of human moral reasoning with foundations such as care/harm, fairness/cheating, loyalty/betrayal, authority/subversion, and purity/degradation. We focus on comments that are more related to moral foundations.
%;comments containing the redditor's moral concepts are useful in aligning them with moral rules-of-thumb. 

%For comments in each cluster, 
We compute each comment's moral foundations score by counting the number of word matches with the Moral Foundation Dictionary \citep{frimer2019moral}, which has approximately 2,000 words with their corresponding moral foundations. For each cluster, we keep the top 6 comments with the largest moral foundations score, resulting in 120 comments in total for each redditor. The major reason to keep the small number of the comments is because there are users who commented very little to specific clusters, and setting the same fixed number of comments is more efficient in model training. Examples are illustrated in Table \ref{tab:mg-example}. %DG: cite MFT + motivate its use (align with higher level RoT?)

Rather than equally considering all subjective ground comments in the cluster, a separate attention module is trained using $\mathcal{D}^+$. The module computes the attention score between the redditor's subjective ground and situation representations, and is followed by feed-forward networks to predict moral judgments. We assume that this module learns weights over subjective ground comments when they are conditioned to the input situation. In other words, we expect the attention layer to highlight the redditor's subjective ground comments that are most relevant to predict correct moral judgments on the given situation.

\subsection{Value Attention Training Module}
\citet{schwartz1992universals} introduced a theory of basic human values to characterize cultural groups, societies, and individuals. In this study, values are used to explain an individual's motivational bases of certain behavior. Value attention training module aims to map individual's subjective ground comments to more abstract values---in our case, moral rules-of-thumb in $\mathcal{D}$ can be used as value candidates. This process is essential because situation clusters capture broad topics rather than fine-grained talking points, thus the redditor's subjective ground comments might not be directly applied to an input situation. For instance, clusters regarding romantic relationships, family members, or kids tend to vary a great extent, and it is challenging to acquire a fixed number of comments that could cover all situations in the cluster.
% To make the model utilize a set of values that are directly related to the given situation, we first obtain a value candidate matrix that is composed of a fixed number of maxims---some of them support the acceptability of the situation, while others do the opposite. In our case, moral rules-of-thumb in $\mathcal{D}$ can be used as value candidates. 

In value attention training, we compute attention scores between value candidates and the subjective ground of the redditors which has already been trained in the previous module. Assuming that the subjective ground has high correlation with moral judgments of the situation, the rule-of-thumb that has the highest attention weight would be highly correlated to the judgment as well. This module projects one's subjective ground on the value candidates to assess the given situation.
%The attention scores represent the vector similarity between each element of the subjective ground and rules-of-thumb.

After computing attention weights among the value candidates, we obtain a weighted sum and consider it as the representation of the value that would speak for the redditor. The final feed-forward classifier layer takes the weighted sum of values and the input situation representation, concatenates them, and predicts the redditor's moral judgments on the situation.

\section{Experiments}

We implement models varying in attention structures and subjective ground representations. Macro F1 is used as an evaluation metric for accuracy since the label distribution is unbalanced and the two classes are equally important. Implementation details are described in Appendix \ref{sec:appendix-implementation-details}. Codes are released for future reference.\footnote{https://github.com/younggns/subjective-ground}

\subsection{Baseline}
Baseline models are implemented to measure the difficulty of identifying a redditor's judgment pattern on given situations without providing context information. We define Transformer-based sequence classifiers for each redditor and train the models to predict the redditor's judgments given the input situations.

Observing the limited amount of training instances in $\mathcal{D}$ per redditor, we make use of a larger dataset, $\mathcal{D}^+$ to fine-tune the encoder layer. We first train sequence classifiers with the same objectives using $\mathcal{D}^+$ and share the encoder for fitting the model to $\mathcal{D}$. This model is denoted as Baseline, fine-tuned encoder.
%$\mathcal{D}^+$ contains approximately 4.6 times more number of situation-judgment pairs for the most active redditors compared to $\mathcal{D}$. 

\subsection{Rules-of-Thumb Self Attention}
As a part of ablation studies, we implement a model that predicts redditors' judgments with the help of rule-of-thumb candidates of input situations. The major difference between this model and our proposed model is the use of the redditor subjective ground; this model does not take subjective ground into consideration. We compute the self attention of rule-of-thumb representations and concatenate with input representations to predict the judgments.
%, hence there is no Key matrix to put into the attention layer
% We consider the rule-of-thumb representations as a Key matrix, making the value attention layer compute the self attention of rules-of-thumb.

\subsection{Subjective Ground Models}
Our model learns the correlation between the input situations and the redditor's subjective ground, and later identifies the most relevant value to the subjective ground for predicting moral judgments. We denote our model as Subjective Ground Attention. 

To investigate the effect of subjective ground attention layers, we implement the Static Subjective Ground model; this model uses the exact same structure as Subjective Ground Attention, without assigning or learning any attention weights. This model therefore takes all subjective ground comments equally.%same subjective ground comment representations without assigning any learned weights to them.

Another variation, Subjective Ground Attention w/o RoT, is a model that uses attention-weighted subjective ground comments without integrating moral rules-of-thumb. This model evaluates the efficacy of mapping subjective ground comments to rule-of-thumb candidates that are directly related to input situations.

One may argue that simply adding more layers and parameters could help improve the performance regardless of the learning aspects of the model. Thus to analyze the efficacy of using a redditor's previous comments as subjective ground, we put a randomly initialized matrix as the subjective ground. This model is denoted as Latent Subjective Ground.

\begin{table}\footnotesize
\centering
\begin{tabularx}{0.93\columnwidth}{l|c}
\hline
\textbf{Model} &  \textbf{F1 (stdev)}\\
\hline
Random Prediction & 48.77 (0.60)\\
\hline
Baseline & 58.61 (0.97)\\
Baseline, fine-tuned encoder & 59.12 (0.34)\\
\hline
Rules-of-Thumb Self Attention & 59.66 (0.72) \\
\hline
Latent Subjective Ground & 59.16 (0.67) \\
Static Subjective Ground & 60.15 (0.73)\\
Subjective Ground Attention w/o RoT & 60.83 (0.61) \\
Subjective Ground Attention & \textbf{61.05 (0.21)}\\
\hline
\end{tabularx}
\caption{\label{tab:experimental-results}
F1 measures of the models in predicting moral judgments.
}
\end{table}

\section{Discussion}
In this section, we analyze the prediction results of different models and discuss the effectiveness of each model component. Additionally, we perform a qualitative analysis of subjectivity explanations of the model.

\subsection{Prediction Accuracy}
Table \ref{tab:experimental-results} reports the average macro F1 scores and standard deviation of five experiments of each model. The overall macro F1 scores of the implemented models are not high. We suppose the task is naturally challenging because the number of training instances is insufficient to learn the moral judgment patterns of input situations---there are less than 500 instances for each redditor on average. Using more data to fine-tune the encoder slightly influences the performance on $\mathcal{D}$---the baseline model with its Transformer encoder fine-tuned with $\mathcal{D}^+$ shows higher accuracy compared to the baseline model. 
%of high similarities among input situations---without considering the content of the Reddit posts, situation descriptions in the dataset are very similar to one another, yet redditors provide different judgments on them. Another reason is because 
% Baseline F1 scores have in fact improved to 62.96 when the model was trained and tested on $\mathcal{D}^+$.
%, which is approximately five times larger than $\mathcal{D}$. 

Using more context information such as rule-of-thumb candidates and subjective ground comments improve the model accuracy to a certain extent. Both Rules-of-Thumb Self Attention and Static Subjective Ground models perform better while Latent Subjective Ground model does not. This shows that the redditor's previous comments help understand their subjectivity. Subjective Ground Attention, our proposed model, is the most efficient way of integrating both rules-of-thumb and subjective ground comments. The prediction accuracy of Subjective Ground Attention is more improved than that of the models without subjective ground comments in a statistically significant way, showing p-value less than $0.01$.
%This shows that over-parameterizing the model will not guarantee improvement, especially if the number of training instances is small for each model. 
%The performance gap between Rules-of-Thumb Self Attention and Static Subjective Ground implies that taking a redditor's previous activities into consideration make better use of rule-of-thumb candidates in moral judgment predictions. 

\begin{figure}
    \centering
    \includegraphics[width=0.47\textwidth]{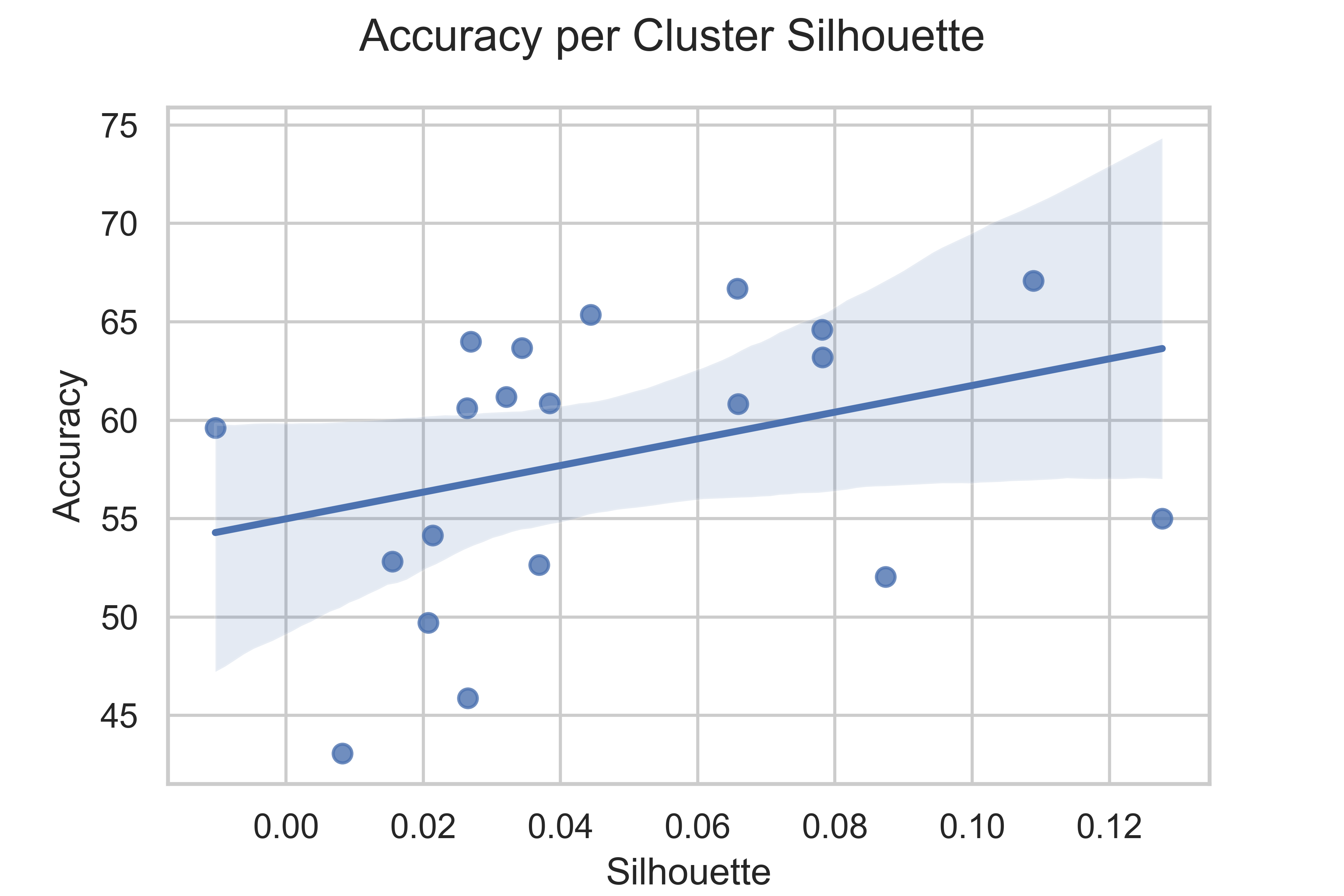}
    \caption{Prediction accuracy of different clusters with respect to silhouette scores. Each dot in the graph represents a distinct cluster.}
    \label{fig:cluster-analysis}
\end{figure}

We further break down the outputs and analyze prediction accuracy for each cluster, assuming that the difficulty of prediction varies based on the topic and the quality of clustering. The gap between the highest and the lowest accuracy cluster is 24\% which supports our assumption. To identify the attributes that are correlated to the cluster accuracy, we investigate a few attributes such as cluster size, intra-cluster distances, and label distributions. 

Figure \ref{fig:cluster-analysis} shows cluster accuracy based on their silhouette score, which is high when items in a cluster are close together and distant from other clusters. There are a few outliers showing high accuracy with low silhouette and vice versa, yet the graph shows positive correlations with Pearson's correlation coefficient of 0.34. This implies that more well-clustered and semantically distinctive situations tend to provide better accuracy. Detailed descriptions of each cluster and their predictions are described in Appendix \ref{sec:appendix-cluster-accuracy}.
%The outliers that have low accuracy with high silhouette scores are situations related to pets and roommates, and the outlier with high accuracy with low silhouette score is situations related to saying something offensive to someone. The former situation (i.e. high silhouette outliers) are clustered well with a strong keyword (e.g. dogs, roommates) but is difficult to judge them, while the latter has more diverse talking points, but is easier to judge. 

\subsection{Consistency in Attention Weights}
We define a new metric for evaluating the consistency of the attention modules on test data. One of the desired behaviors of the model is its consistency; the rule-of-thumb with the largest attention weight needs to be consistent with the model's prediction. For instance, if the most attended rule-of-thumb supports the acceptability of the input situation, we want the model prediction to be `acceptable' regardless of the ground truth. We manually annotate the acceptability label of 500 rules-of-thumb. Value consistency is then defined as a portion of instances where the highest weighted rule-of-thumb's acceptability label matches the model's final prediction.

\begin{figure*}[!ht]
    \centering
    \includegraphics[width=0.95\textwidth]{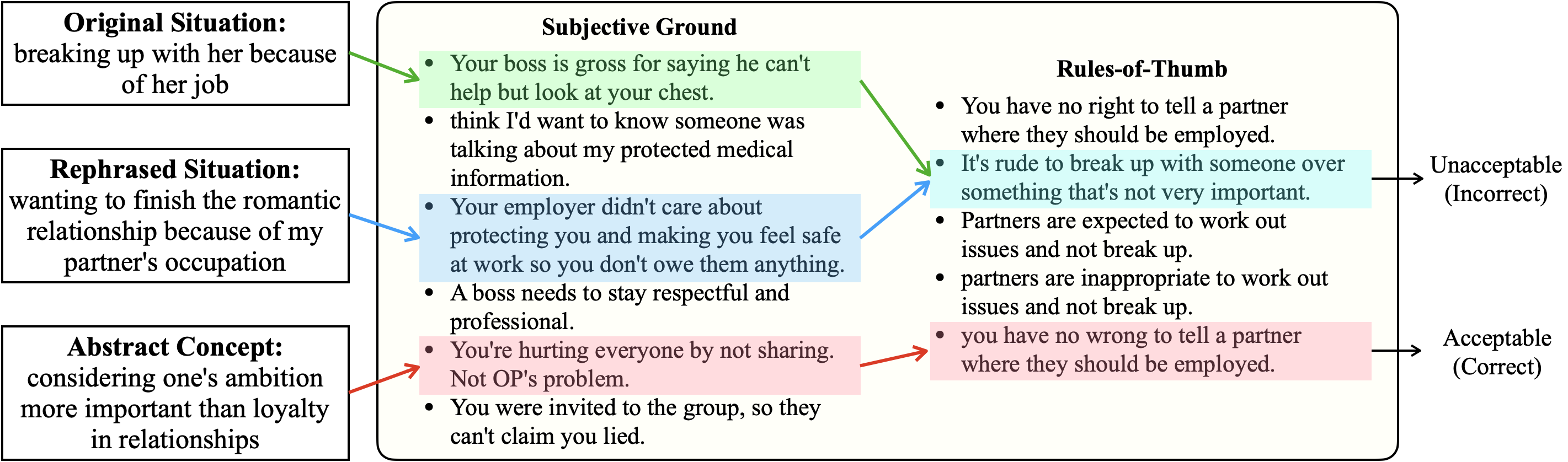}
    \caption{Attention weight flows of the original situations and perturbed inputs. When an abstract concept is given, the model attends to the redditor's subjective ground differently and predicts the judgment correctly.}
    \label{fig:abstract-concept-example}
\end{figure*}

\begin{table}\footnotesize
\centering
\begin{tabularx}{0.877\columnwidth}{l|c|c}
\hline
\multicolumn{3}{c}{\textbf{Attention Consistency Test}}\\
\hline
\multirow{2}{*}{\textbf{Model}}&\multicolumn{2}{c}{\textbf{Consistency}}\\\cline{2-3}

    &\textbf{Value}&\textbf{SG}\\
\hline
Rules-of-Thumb Self Attention & 35.94 & N/A\\
\hline
Static Subjective Ground & \textbf{65.63} & 71.15\\
Latent Subjective Ground & 42.19 & 67.96\\
Subjective Ground Attention & 56.25 & \textbf{72.32}\\
\hline
\multicolumn{3}{c}{}\\
\multicolumn{3}{c}{}\\
\hline

\multicolumn{3}{c}{\textbf{Input Perturbation Test}}\\
\hline
\textbf{Data} &  \multicolumn{2}{c}{\textbf{Accuracy}}\\
\hline
Original Situation & \multicolumn{2}{c}{51.07}  \\
Altered Gender & \multicolumn{2}{c}{45.15} \\
Rephrased Situation & \multicolumn{2}{c}{46.73} \\
% & N/A\\
Abstract Concept & \multicolumn{2}{c}{\textbf{58.18}} \\
% & 33.33 \\
\hline
\end{tabularx}
\caption{\label{tab:discussion-results}
Quality evaluation of attention weights. The upper table reports the consistency measure of the value attentions (\textbf{Value}) and the subjective ground attentions (\textbf{SG}). Note that subjective ground consistency can't be measured in Rules-of-Thumb Self Attention because this model doesn't refer to subjective ground comments. The lower table shows the accuracy of Subjective Ground Attention model on modified inputs.
}
\end{table}

It is more challenging to annotate the acceptability label of subjective ground comments. Thus we design another test, input perturbation, to measure the consistency of subjective ground attentions. The redditor's subjective ground needs to account for situations at inference, and we expect the model to behave consistently for similar situations. From this intuition, we manually create situations that are similar to the original reddit posts.\footnote{Examples are described in Appendix \ref{sec:appendix-input-modification}} We apply three levels of similarity; (1) situations where pronouns and gender-specific nouns are altered (e.g. not respecting my mom $\rightarrow$ not respecting my dad), (2) rephrased situations (e.g. not respecting $\rightarrow$ being mean to), (3) abstract concept of the situations where it can be applied to other situations (e.g. revealing someone's secret $\rightarrow$ honesty is more important than relationships). Subjective ground consistency is defined as the portion of modified inputs that have the same attention weight order as the original input. Evaluation results are described in Table \ref{tab:discussion-results}.

\begin{table*}[!ht]
    \centering
    \includegraphics[width=0.95\textwidth]{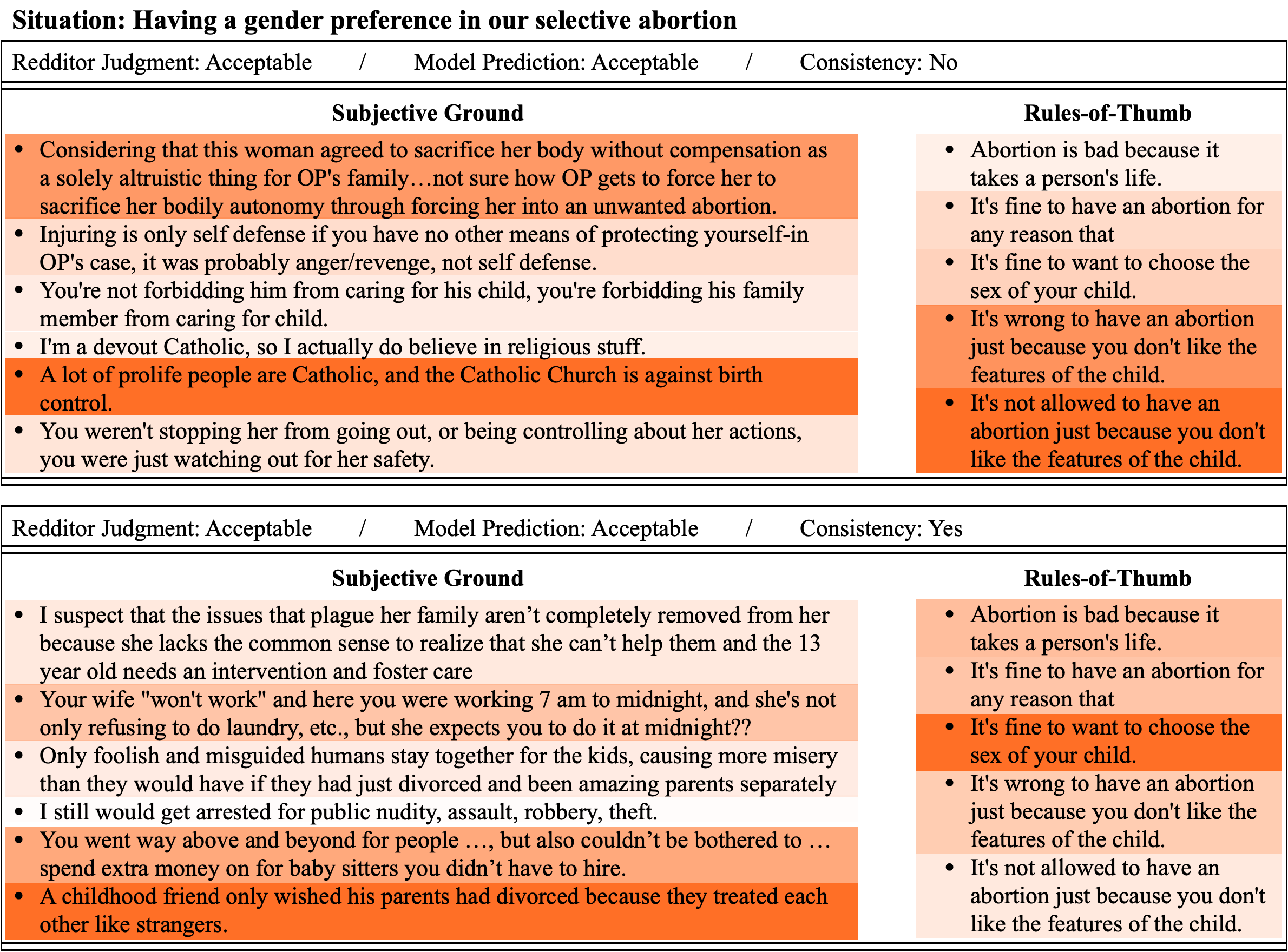}
    \caption{Attention weights among subjective ground comments and rule-of-thumb candidates given an input situation. The two redditors have the same judgments yet they differ in the rules-of-thumb attention and their consistency.}
    \label{tab:analysis-example}
\end{table*}

The value consistency of Rules-of-Thumb Self Attention and Latent Subjective Ground models are surprisingly low, implying these models learn rules-of-thumb attention weights without their actual relatedness to the moral judgments---right for the wrong reasons. We further investigate the reason and observe that the model tends to give high weights on some specific rules-of-thumb, possibly texts that are more familiar to the pre-trained Transformers regardless of the redditor's judgments. Static Subjective Ground model gives the highest consistency score, confirming the efficacy of using redditor's subjective ground comments. This model exceeds the value consistency measure of our proposed model, Subjective Ground Attention, suggesting rules-of-thumb attentions become more consistent with the model's final prediction when using all subjective ground comments equally. Our proposed model achieves the highest score in subjective ground consistency tests. This implies that Subjective Ground Attention learns consistent attention weights with respect to similar inputs.
%Another observation is that the consistency varies substantially for different redditors---some redditors show more than 80\% consistency whereas others score 50\%. This implicates using previous activities as one's subjective ground is reasonable depending on whether the redditors share their thought processes in their comments.

Another interesting finding is that our model gives more accurate predictions on the abstract concept inputs; when the model is conditioned to the abstract concepts, the order of the subjective ground attention weights changes and it leads to better prediction results. Referring to an example illustrated in Figure \ref{fig:abstract-concept-example}, the model attends more on the first and the third comments when the original situation and the rephrased situation is given. This is because the model picks up keywords like job and occupation, and considers the situation as a job/work related issue. For an abstract concept input, however, the fifth subjective ground is attended the most as the model now sees the situation as one regarding relationships. The different weights over subjective ground affects the attention weights on rules-of-thumb, hence they impact the final model prediction.
These results suggest that our proposed model learns one's general perspectives on morally competing values although applying this knowledge to specific situations is yet challenging.

\subsection{Qualitative Analysis of Subjective Ground}
In this section, we qualitatively analyze the redditors' subjective ground attention and its relatedness to rules-of-thumb attention.

We illustrate a case where two redditors comment on the same post in Table \ref{tab:analysis-example}. The upper case is where the model prediction is not consistent with the most attended rule-of-thumb. We observe that the model gives higher weights to abortion-related subjective ground comments, implying that the redditor would consider the given situation unacceptable. The value attention module chooses the last rule-of-thumb which opposes the idea of abortion, showing the consistency between subjective ground and value attention. However, the final prediction of the model is `acceptable', suggesting that the classifier does not use the weighted rule-of-thumb representations correctly. This analysis matches suboptimal value consistency results of Subjective Ground Attention in Table \ref{tab:discussion-results} and raises the necessity of developing classifiers that can better understand value attentions with respect to moral judgments.

The next redditor, on the other hand, does not include any abortion-related comments as subjective ground. The model attends to the last subjective ground comment that contains keywords related to family and their separation---divorced, strangers. This example highlights the case where the topic distribution of a redditor's subjective ground is not comprehensive enough with respect to the given situation. In such case, the attention module focuses on the ground that is potentially associated with the situation and give high attention weights on the related rules-of-thumb. We anticipate that the model will be more accurate and consistent using subjective ground that is clustered based on more fine-grained talking points.

Overall, we observe the consistency in subjective ground and value attentions. We expect the model's prediction accuracy and the quality of explanations can be further improved using more fine-grained activities of individuals and a neural component that can better learn the correlation between rules-of-thumb and moral judgments.

\section{Related Work}
\textbf{Explainable AI} \quad As deep neural language models improve the accuracy of many different downstream NLP tasks, measuring the accountability and interpretability of these models has been recently gaining interest in the research community.

Local explanation methods aim to provide rationales of the model in predicting a specific input. Recent neural models majorly explain the model behaviors by visualizing the saliency map of the first derivatives of the encoder \citep{ross2017right, wallace-etal-2018-interpreting}, attention layers \citep{xie-etal-2017-interpretable, mullenbach-etal-2018-explainable}, perturbing inputs \citep{sydorova-etal-2019-interpretable}, and applying rules and templates \citep{abujabal-etal-2017-quint, pezeshkpour-etal-2019-investigating}. \citet{rajani-etal-2019-explain} collect human explanations for commonsense reasoning in the form of text, and train language models to generate the explanations given pairs of commonsense questions and answers. \citet{aubakirova-bansal-2016-interpreting} investigate how neural network models predict the politeness of input text by visualizing activation clusters, saliency heatmaps of the first derivatives, and word representation transformations in the embedding space. \citet{ribeiro2016should} proposes a framework where an interpretable model, trained to minimize the distance to the classifier predictions, explains the model prediction with absence/presence of specific words.

\textbf{Perspective Identification} \quad Identifying perspectives from the text has been steadily studied in many sub-fields of NLP. \citet{greene-resnik-2009-words} defines perspectives as an individual's syntactic packaging of the information and analyzes different usage of linguistic cues in articles. \citet{choi-wiebe-2014-effectwordnet} adopts a simple symbolic relation, positive and negative connotations towards events and concepts, to the existing WordNet \citep{miller1990introduction} hierarchy for inferring the point of view of an opinion. There are more complex concepts and structures---analyzing political framing and agenda-setting \citep{field-etal-2018-framing, roy-goldwasser-2021-analysis}, encoding political perspective flows in social settings via Graph Convolutional Network \citep{li-goldwasser-2019-encoding}---have been studied.

This research paper is positioned in the intersection of explainable AI and perspective identification. We examine several models that can approximate an individual's subjective ground in a human-readable way, as well as interact with diverse daily situations to infer individual's perspectives on the author and their situations.

\section{Conclusion}
This paper proposes a neural model, Subjective Ground Attention, that represents an individual's subjective preference with their previous activities and explains the reasoning behind their moral judgments on diverse social situations by spotlighting the most relevant subjective ground. We explore situations posted on a Reddit community, \verb|r/AmITheAsshole|, and analyze active redditors' judgments on these situations indicating whether or not the author's behavior is acceptable. Upon attending subjective comments and moral rules-of-thumb, experimental results show that the model provides reasonable explanations without sacrificing prediction accuracy. Although attention weights on moral rules-of-thumb show suboptimal consistency with the model's prediction, we illustrate the model's consistency in attention weights on subjective ground comments. We further claim that our model better captures one's perspectives on abstract moral concepts.
%improves both the prediction accuracy and the explainability of individual's subjectivity. 

\section*{Limitations}
One of the major limitations of this work is the absence of Reddit post contents. Although reading the content of the post is crucial in fully comprehending and judging the situation, we decide not to include the content mainly because of the size of training instances. A large volume of the text in the post content would have hindered the model from good generalization.

Another shortcoming is the subjectivity explanation tool---moral foundation annotations. Ideally moral rules-of-thumb represent an individual's biases in judging situations, yet in reality the annotated rules-of-thumb do not cover all types of biases related to the situation. Additionally, many of the manually crafted rules-of-thumb will not help the model learn different types of biases since they are largely similar to the original rules-of-thumb.

Lastly, individual subject ground modeling in this work is over-simplified. We construct the redditors' subjective ground solely based on their previous comments in the same subreddit, and there is more context information that could potentially help analyze an individual, such as the posts submitted by the redditor and information about subreddits they are actively involved. Rather than choosing the subjective ground comments based on the word matches with Moral Foundation Dictionary, applying more sophisticated methods in identifying moral foundations, such as moral foundations framing \citep{roy-goldwasser-2021-analysis}, would also lead to a better subjective ground. In analyzing a different domain in future work, we could also take an individual's identity---demographic, social, political---into consideration when modeling subjective ground.

\section*{Ethics Statement}
To the best of our knowledge, this work has not violated any code of ethics. We anonymize redditor information in the paper as well as in the datasets we share to the public. This paper shows different redditors' subjective ground, yet there is no discrimination in choosing redditors of the interest; the redditors are selected solely based on the number of comments they have left on this subreddit. We provide the code and datasets for future reproducibility of the work.

\section*{Acknowledgements}
This project was partially funded by  NSF IIS-2048001 and DARPA CCU program. The views are the authors' and should not be interpreted as necessarily representing the official policies, either expressed or implied, of DARPA, or the U.S. Government.

\bibliography{anthology,custom}
\bibliographystyle{acl_natbib}

\appendix

\section{Experiment Details}
\label{sec:appendix-exp}
% \begin{itemize}
%     \item description of math, algorithms, models
%     \item submit codes, dependency specifications
%     \item computing infrastructures
%     \item average runtime (train, inference)
%     \item num parameters 
%     \item valid performance for test results
%     \item evaluation metrics
%     \item num of training / eval runs
%     \item bounds for hyperparameters
%     \item hyperparameter configurations
%     \item num of hyperparameter search trials
%     \item method of choosing hyperparameters (manual, uniform sample), criterion for selection (accuracy)
%     \item summary statistics (mean, var, error bars)
% \end{itemize}
\subsection{Model Formalization}
\label{sec:model-formalization}
While training the subjective ground module of the redditors, we make use of the input situations $X^{+} = \{x_{1}^+, x_{2^+}, \ ... \ , x_{N}^+\} \in \mathcal{D}^+$, where $M$ redditors in $\{u_1, u_2, \ ... \ , u_M\}$ have commented on. An $i$-th redditor $u_i$ is represented as a set of $G$ subjective ground comments, $u_i = \{c_{i1}, \ ... \ , c_{iG}\}$. Binary output labels, $Y^+$, indicating acceptable / unacceptable judgments on each input situation can be denoted as $Y^+ = \{0, 1\}^{L}$ where $L$ is the total number of instances. Note that $L \neq NM$ since not all redditors commented on all input situations in the dataset.

We use another dataset with moral rules-of-thumb annotation, $\mathcal{D}$, for training and testing the model. Similar to $\mathcal{D}^+$, input situations and output labels are denoted as $X = \{x_{1}, x_{2}, \ ... \ , x_{n}\}$, and $Y = \{0, 1\}^{l}$, where $n$ and $l$ are the number of situations and instances in $\mathcal{D}$, respectively. Additionally, there are $K$ rules-of-thumb annotated for each situation. An input situation $x_{j}$ is mapped to the rule-of-thumb candidates, $V_j = \{v_{j1}, \ ... \ , v_{jK}\}$.

The first step of training is on the redditors' subjective ground using $\mathcal{D}^+$. Using a pre-trained Transformer encoder \citep{wolf2019huggingface}, we represent an $i$-th redditor's subjective ground as $\texttt{SG}_{i} \in \mathbb{R}^{G \times h}$ where each row of the matrix is the encoded subjective ground comment and $h$ is the encoder dimension. Assuming this redditor has commented on the $j$-th input situation in $\mathcal{D}^+$, encoded as $z_{j}^+$, the subjective ground training module is designed as follows:

\par
{\footnotesize
\begin{align*}
    z_{j}^+ &\leftarrow \textrm{Transformer}(x_{j^+})
    \\
    \texttt{SG}_{i} &\leftarrow \textrm{Transformer}(\{c_{i1}, \ ... \ , c_{iG}\})
    \\
    a_{i, j}^+&=\textrm{Multihead}(\texttt{SG}_{i}, z_{j}^+, \texttt{SG}_{i}) \\
    \hat{y} &= W_{CLF}[\Sigma a_{i,j}^+\texttt{SG}_{i} \ ; \ z_{j}^+]\\
    \mathcal{L} &= \textrm{CrossEntropy}(y, \hat{y})
\end{align*}
}%

\par
We follow the basic structure of the Multi-head attention proposed by \citet{vaswani2017attention}, where different representations of attention inputs are combined:
\par
{\footnotesize
\begin{align*}
    \textrm{Multihead}(Q, K, V) &= [\textrm{head}_1 ; \ ... \ ; \textrm{head}_h]W^O\\
    \textrm{s.t. head}_i &= f \left( \frac{QW_i^Q(KW_i^K)^\texttt{T}}{\sqrt{d_k}}VW_i^V \right) \\
    \textrm{where } f(\cdot) &: \textrm{ softmax function }\\
    \textrm{and }d_k &: \textrm{ model dimension}\\
    W_i^Q, W_i^K, W_i^V &: \textrm{ input projections}
\end{align*}
}%
\par

After subjective ground is trained, we learn the attention between the redditor's subjective ground and moral rules-of-thumb of the situations in $\mathcal{D}$. We use  the encoder that is fine-tuned in the previous step and obtain an encoded value candidate representation of a $j$-th situation $x_j$ as $\texttt{VAL}_j \in \mathbb{R}^{K \times h}$ where each row of the matrix is the encoded rule-of-thumb. Suppose the $i$-th redditor commented on $x_j$, the model learns the data as follows: 
\par
{\footnotesize
\begin{align*}
    z_{j} &\leftarrow \textrm{Transformer}(x_{j})
    \\
    \texttt{SG}_{i} &\leftarrow \textrm{Transformer}(\{c_{i1}, \ ... \ , c_{iG}\})
    \\
    \texttt{VAL}_{j} &\leftarrow \textrm{Transformer}(\{v_{j1}, \ ... \ , v_{jK}\})
    \\
    a_{i, j}^{\texttt{SG}} &=\textrm{Multihead}(\texttt{SG}_{i}, z_{j}, \texttt{SG}_{i}) \\
    a_{i, j}^{\texttt{VAL}}&=\textrm{Multihead}(\texttt{VAL}_{j},  a_{i,j}^{\texttt{SG}}\texttt{SG}_{i}, \texttt{VAL}_{j}) \\
    \hat{y} &= W_{CLF}[\Sigma a_{i,j}^{\texttt{VAL}}\texttt{VAL}_{j} \ ; \ z_{j}]\\
    \mathcal{L} &= \textrm{CrossEntropy}(y, \hat{y})
\end{align*}
}%
\par

% \begin{table}\scriptsize
% \centering
% \begin{tabularx}{0.9\columnwidth}{l|c|c}
% \hline
% \textbf{Model} &  \textbf{\# params} &  \textbf{time}\\
% \hline
% Random Prediction & N/A & N/A\\
% \hline
% Baseline & 18M & 00:20:00\\
% Baseline, fine-tuned encoder & 18M & 00:20:00\\
% \hline
% Rules-of-Thumb Self Attention & 115M & 1:00:00 \\
% \hline
% Latent Subjective Ground & 115M & 1:20:00\\
% Static Subjective Ground & 115M & 1:30:00\\
% Subjective Ground Attention w/o RoT & 115M & 1:30:00 \\
% Subjective Ground Attention & 115M & 1:30:00\\
% \hline
% \end{tabularx}
% \caption{\label{appendix-num-params}
% Number of parameters and training time for each model. We excluded the number of parameters in the Transformer encoder that is fine-tuned. We also excluded the number of parameters and training time for training subjective ground attention modules since this step is applied to all non-baseline models.
% }
% \end{table}

% \begin{table}\footnotesize
% \centering
% \begin{tabularx}{0.82\columnwidth}{l|c}
% \hline
% \textbf{Model} &  \textbf{F1}\\
% \hline
% Latent Subjective Ground & 85.03 \\
% Static Subjective Ground & 97.32\\
% Subjective Ground Attention w/o RoT &  96.81\\
% Subjective Ground Attention & 95.17\\
% \hline
% \end{tabularx}
% \caption{\label{experimental-results-valid}
% F1 measures of the proposed models on validation sets.
% }
% \end{table}

\subsection{Implementation Details}
\label{sec:appendix-implementation-details}
We use a pre-trained DistilBERT-base-uncased for the text encoder, distributed by \citet{wolf2019huggingface}. For attention layers, we implement multi-head scaled dot-product attention layers with 12 heads. Classifiers are two-layer linear networks followed by Cross Entropy loss and Adam optimizer \citep{kingma2014adam} with static learning rates. The final model considers $d_k$ as 1 in the multi-head attention layers, because normalizing attention weights by the square root of the model dimension generates more equally distributed attention weights over subjective ground comments and rules-of-thumb. All experimental results are the average of five separate runs.

Our models are trained and tested using NVIDIA Tesla V100 GPU and the average time for training the full model is approximately 6 hours, while the training time for the baseline models are 20 minutes. %The number of parameters for the largest model is $30\ \text{redditors} \times (\text{DistilBERT parameters} + 768*768*4 \ (\text{two attention modules}) + \text{two FC classifiers})$, which is about 3M parameters without considering language model parameters, for each redditor. 
% The approximate training time and the number of parameters for other models are described in Table \ref{appendix-num-params}.

We manually select the hyperparameters to tune---the number of attention heads and learning rates. The selection criterion for the hyperparameters was the average F1 score of five experiments on test data. We set the possible number of heads either 1 or 12, where 1 means single attention head. Hyperparameters are searched using grid search, in the boundary from 1e-6 to 1e-3. We also implemented learning rate warm-ups, where the learning rate increases for the first few steps, then it decreases logarithmically.

\subsection{Cluster Results}
\label{sec:appendix-cluster-accuracy}
We break down the model performance based on each cluster. The F1 score of each cluster and the most representative situations in the cluster are dsecribed in Table \ref{appendix-cluster-results}.

\begin{table*}\footnotesize
\centering
\begin{tabularx}{\textwidth}{X|X}
\hline
\textbf{Cluster 0}: kids, \textbf{Accuracy}: 52.81 &\textbf{Cluster 1}: romantic-relationships-GF, \textbf{Accuracy}: 63.99 \\ 
\hline
telling my son's stepmom I don't care about her kid& telling my girlfriend to stop talking to a girl\\
refusing to babysit our daughter& telling my gf she's over reacting\\
not letting in laws see our child& accusing my girlfriend of wanting to cheat\\
\hline
\hline
\textbf{Cluster 2}: not wanting, \textbf{Accuracy}: 45.89 &\textbf{Cluster 3}: romantic-relationships-BF, \textbf{Accuracy}: 54.14 \\ 
\hline
not wanting to go to my boyfriends moms house& not telling my boyfriend about an ex\\
not wanting to spend time with my mother&telling my boyfriend not to contact me again \\
not wanting my stepmom to meet my boyfriend again& refusing to see a comprise with my boyfriend\\
\hline
\hline
\textbf{Cluster 4}: telling, \textbf{Accuracy}: 61.19&\textbf{Cluster 5}: siblings, \textbf{Accuracy}: 66.69 \\ 
\hline
telling my dad that his wife isn't family& telling my sister she's not family\\
telling my daughter to leave my husband alone&Telling My Brother Not To Come To My Sister's Wedding \\
telling my girlfriend to back off from my daughter& asking my parents to disinherit my half sister\\
\hline
\hline
\textbf{Cluster 6}: romantic-relationships, \textbf{Accuracy}: 63.69&\textbf{Cluster 7}: wedding, \textbf{Accuracy}: 67.08 \\ 
\hline
not letting my girlfriend into my parents house& not inviting my sister-in-law to my wedding\\
giving my girlfriend an ultimatum regarding their best friend& not going to my sister’s wedding after being initially uninvited\\
not telling my boyfriend that his friend made a pass at me& not letting my future in-laws invite people to our wedding\\
\hline
\hline
\textbf{Cluster 8}: cars, \textbf{Accuracy}: 52.65&\textbf{Cluster 9}: kids, \textbf{Accuracy}: 43.08 \\ 
\hline
not giving my neighbor a parking spot& leaving my son behind on our family vacation\\
telling someone not to come up to my car& dropping my stepdaughter off at her mothers house\\
calling the police on someone who parked in my driveway& letting my daughter 'take over' my son's birthday present\\
\hline
\hline
\textbf{Cluster 10}: roommates, \textbf{Accuracy}: 52.04&\textbf{Cluster 11}: emotional-burst, \textbf{Accuracy}:  63.20\\ 
\hline
not wanting my roommate to move out& being upset at my mom\\
asking my roommate's significant other to move out& being mad at my significant other\\
refusing to switch bedrooms with my future roommate& being pissed off at my girlfriend\\
\hline
\hline
\textbf{Cluster 12}: refusing-family-behaviors, \textbf{Accuracy}: 65.35&\textbf{Cluster 13}: money, \textbf{Accuracy}: 60.62 \\ 
\hline
not letting my daughter go to my in-laws house& not wanting to pay for something my Girlfriend offered to pay for\\
not allowing my mom to live with me& wanting my co-worker to pay me back\\
refusing to bring my little brother to birthdays& Asking my Fiancee to Pay 1/3 of the Bills\\
\hline
\hline
\textbf{Cluster 14}: coworkers, \textbf{Accuracy}: 60.87&\textbf{Cluster 15}: pets, \textbf{Accuracy}: 55.01 \\ 
\hline
not telling my co-worker that he's going to be fired& not giving my dog back to his original owner\\
getting a co-worker fired for something I also did& telling my wife she can't get rid of the dog\\
Refusing To Work With My Coworker& not letting a homeless guy pet my dog\\
\hline
\hline
\textbf{Cluster 16}: offensive-behavior, \textbf{Accuracy}: 59.62&\textbf{Cluster 17}: parents, \textbf{Accuracy}:  49.71\\ 
\hline
saying something embarrassing about one of my students& telling my mom she didn't raise me\\
making racist and sexist jokes& telling my parents they're out of touch\\
refusing to forgive former classmates for being homophobic& telling my dad's wife her family's not my family\\
\hline
\hline
\textbf{Cluster 18}: food, \textbf{Accuracy}: 64.59&\textbf{Cluster 19}: friendship, \textbf{Accuracy}:  60.83\\ 
\hline
telling my coworker to stop eating my food& telling my friend off\\
not letting my spouse eat my meal prepped food& ruining my friend's relationship\\
not eating the food my coworkers wife made& telling my friend she's out of line\\
\hline

\end{tabularx}
\caption{\label{appendix-cluster-results}
Each cluster's topic, most representative situation examples, and their F1 accuracy on our proposed model.
}
\end{table*}

\begin{table*}\footnotesize
\centering
\begin{tabularx}{\textwidth}{X|X}
\hline
\textbf{Original}: telling my wife that when/if she quits her job to be a SAHM I am not planing to do much in the way of household chores anymore &  \textbf{Original}: ``being triggered'' by my boyfriend setting rules for my pregnancy weight gain\\
\textbf{Gender}: telling my husband that when/if he quits his job to be a SAHD I am not planing to do much in the way of household chores anymore &  \textbf{Gender}: ``being triggered'' by my girlfriend setting rules for my weight gain\\
\textbf{Rephrase}: suggesting my wife to balance household chores since I will be working while she quits her job &  \textbf{Rephrase}: being upset at my boyfriend when he plans to prevent me from eating too much during pregnancy\\
\textbf{Abstract}: preferring more fairness over helping out my partner &  \textbf{Abstract}: not wanting to be controlled by my partner's concerns about my health\\
\hline
\textbf{Original}: not going to my girlfriends dads funeral &  \textbf{Original}: having a gender preference in our selective abortion\\
\textbf{Gender}: not going to my boyfriends moms funeral &  \textbf{Gender}: having a racial preference in our selective abortion\\
\textbf{Rephrase}: not wanting to attend to my girlfriends dads funeral &  \textbf{Rephrase}: deciding to have abortion based on the baby's sex\\
\textbf{Abstract}: putting my belief first even my partner has lost their loved ones &  \textbf{Abstract}: believing choice is more important than life\\
\hline
\textbf{Original}: telling DH that I will not let his mom pick her grandmother name &  \textbf{Original}: shaming my sister-in-law because she was mean to me\\
\textbf{Gender}: telling DH that I will not let his dad pick his grandfather name &  \textbf{Gender}: shaming my borther-in-law because he was mean to me\\
\textbf{Rephrase}: not wanting to name my children that my MIL picked &  \textbf{Rephrase}: disrespecting my sister-in-law by making fun of her because she was mean to me\\
\textbf{Abstract}: wanting more freedom in raising kids over respecting the opinion of my parents &  \textbf{Abstract}: revenging someone in the family for their behavior on me\\
\hline
\textbf{Original}: breaking up with him because of his job &  \textbf{Original}: denying my wife a new kitchen\\
\textbf{Gender}: breaking up with her because of her job &  \textbf{Gender}: denying my husband a new kitchen\\
\textbf{Rephrase}: wanting to finish the romantic relationship because of my partner's occupation &  \textbf{Rephrase}: not allowing my wife to get a new kitchen\\
\textbf{Abstract}: considering one's ambition more important than loyalty in relationships &  \textbf{Abstract}: not wanting to waste money on my partner's desire\\
\hline
\textbf{Original}: taking my daughter to get her hair dyed against my wifes wish &  \textbf{Original}: telling my sister's boyfriend the truth about her\\
\textbf{Gender}: taking my son to get his hair dyed against my husbands wish &  \textbf{Gender}: telling my brother's girlfriend the truth about him\\
\textbf{Rephrase}: letting my daughter to get her hair dyed although my wife did not want it &  \textbf{Rephrase}: revealing a big secret about my sister to her boyfriend\\
\textbf{Abstract}: putting my kid's desire first over my partner's thought &  \textbf{Abstract}: considering honesty is always more important even though it would break up the relationships\\
\hline

\end{tabularx}
\caption{\label{appendix-perturbed-situation}
Input situations and their modification for perturbation test.
}
\end{table*}

\section{Datasets}
\label{sec:appendix-datasets}
% \begin{itemize}
%     \item languages, num of instances, label distribution
%     \item Train/valid/test splits
%     \item pre-processing steps (data that were excluded)
%     \item zip file of the data
%     \item data collection process (annotator instructions, quality control)
% \end{itemize}
Train/valid/test splits of $\mathcal{D}$ were provided by the original dataset, \verb|Social Chemistry 101|, and we used the same splits. For the additionally crawled data, $\mathcal{D}^+$, we randomly divided the splits into 80\%, 10\%, 10\%, while excluding all valid and test samples of $\mathcal{D}$ from the training data of $\mathcal{D}^+$.

Additionally annotated data for consistency evaluation, rules-of-thumb consistency and input perturbation consistency, are annotated by the authors.

\subsection{Input Modification for Subjective Ground Consistency}
\label{sec:appendix-input-modification}
The input situations to modify are selected from the test set of $\mathcal{D}$. To compute the subjective ground consistency of more diverse redditors, we sort the test set situations based on the number of redditors participated. We select 10 situations where the redditors have commented the most, resulting in 162 instances in total. The examples of 10 situations with their original situation descriptions, gender altered descriptions, rephrased descriptions, and abstract concept descriptions are in Table \ref{appendix-perturbed-situation}. 
% This evaluation step can be conducted by running \verb|run_user_analysis.py| script in our code submission.

\subsection{Value Attention Consistency}
\label{sec:appendix-value-attn}
Similar to input modification tests, we sort the test set situations based on the number of redditors participated, and select the top 100 situations. The authors annotated the acceptability label of rules-of-thumb of with respect to the situations, resulting in 500 instances in total. 
% This evaluation step can be conducted by running \verb|run_analysis.py| script in our code submission.

\end{document}